\pdfoutput=1

\documentclass[11pt]{article}

\usepackage{acl}

\usepackage{times}
\usepackage{latexsym}

\usepackage[T1]{fontenc}

\usepackage[utf8]{inputenc}

\usepackage{microtype}

\usepackage{graphicx}
\usepackage{textcomp}
\usepackage{xcolor}
\usepackage{multirow}
\usepackage{longtable}
\usepackage{booktabs}
\usepackage{multirow}
\usepackage{tabularx}
\usepackage{comment}
\usepackage{hyperref}
\usepackage{amsmath}

\usepackage{ulem}
\usepackage{subfigure}
\AtBeginDocument{\captionsetup[subtable]{position=b}}

\usepackage{caption}

\newcommand{\method}[0]{universal discriminator~}

\title{A Universal Discriminator for Zero-Shot Generalization}

\author{Haike Xu$^{1}$, Zongyu Lin$^{1}$, Jing Zhou$^{1}$, Yanan Zheng$^{2}$\footnotemark[1], Zhilin Yang$^{134}$\footnotemark[1] \\
$^1$Institute for Interdisciplinary Information Sciences, Tsinghua University \\
$^2$Department of Computer Science and Technology, Tsinghua University \\
$^3$Shanghai Artificial Intelligence Laboratory, $^4$Shanghai Qi Zhi Institute \\
\texttt{haikexu@mit.edu}, 
\texttt{\{zyanan,zhiliny\}@tsinghua.edu.cn} \\}

\begin{document}
\maketitle
\renewcommand{\thefootnote}{\fnsymbol{footnote}}
\footnotetext[1]{Corresponding authors.}

\begin{abstract}
Generative modeling has been the dominant approach for large-scale pretraining and zero-shot generalization. In this work, we challenge this convention by showing that discriminative approaches perform substantially better than generative ones on a large number of NLP tasks. Technically, we train a single discriminator to predict whether a text sample comes from the true data distribution, similar to GANs. Since many NLP tasks can be formulated as selecting from a few options, we use this discriminator to predict the concatenation of input and which option has the highest probability of coming from the true data distribution. 
This simple formulation achieves state-of-the-art zero-shot results on the T0 benchmark, outperforming T0 by 16.0\%, 7.8\%, and 11.5\% respectively on different scales. In the finetuning setting, our approach also achieves new state-of-the-art results on a wide range of NLP tasks, with only 1/4 parameters of previous methods.
Meanwhile, our approach requires minimal prompting efforts, which largely improves robustness and is essential for real-world applications. Furthermore, we also jointly train a generalized UD in combination with generative tasks, which maintains its advantage on discriminative tasks and simultaneously works on generative tasks.

\footnotetext[2]{Our code is available at \href{https://github.com/Rafa-zy/UD}{https://github.com/Rafa-zy/UD}.}

\end{abstract}

\section{Introduction}

Generative modeling has been the dominant approach for large-scale pretraining and zero-shot generalization~\cite{gpt3-paper,artetxe2021efficient,rae2021scaling}. 
Combined with prompts~\cite{gpt3-paper}, most of the natural language processing (NLP) tasks can be formulated into the fill-in-the-blank format and perform generative language modeling.
Based on the unified generative formulation, pretrained models such as GPT-3~\cite{gpt3-paper}, BERT~\cite{devlin2018bert,PET-paper}, T5~\cite{T5-paper}, can perform zero-shot inference on new tasks.

More recent work~\cite{T0-paper} proposed to further pretrain a generative T5~\cite{T5-paper} with multitask prompted datasets and has substantially enhanced the performance of zero-shot generalization. 
In contrast, methods based on discriminative modeling~\cite{devlin2018bert} have not been able to achieve state-of-the-art performance on zero-shot learning. The adoption of discriminative approaches for zero-shot learning has been limited in the literature.

\begin{figure}%
     \centering
     \includegraphics[width=\linewidth]{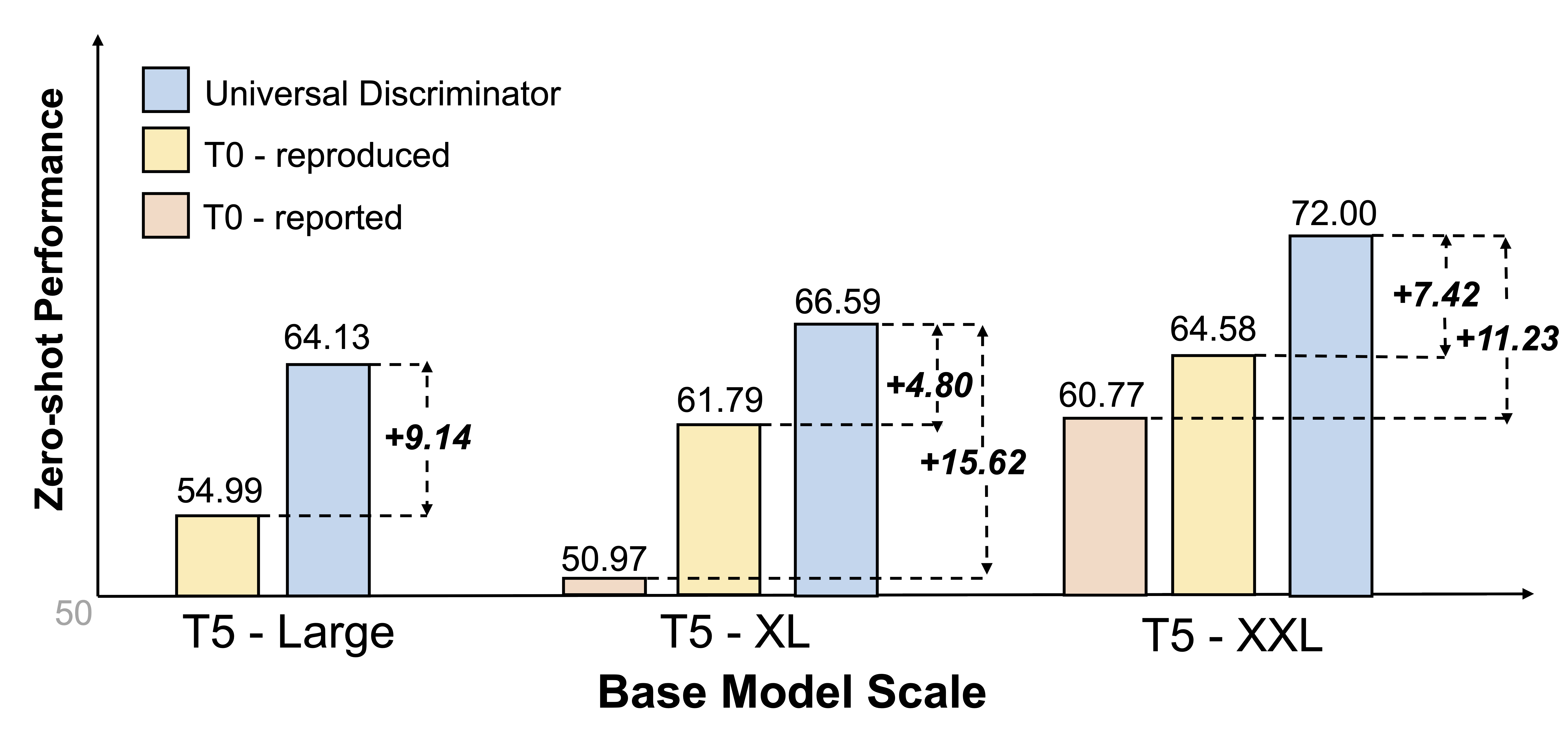}
     \vspace{-15pt}
     \caption{Average zero-shot performance over 11 zero-shot tasks for our Universal Discriminator and T0~\cite{T0-paper}. Our universal discriminator significantly outperforms T0 across three different scales.}
     \label{fig:sota}
     \vspace{-15pt}
 \end{figure}

In this work, we challenge the convention of zero-shot learning and propose to study and improve discriminative approaches. This is motivated by the fact that many NLP tasks can be framed as selecting from a few options; e.g., telling whether sentence A entails sentence B, or predicting which answer is correct for a given question. We call these tasks \textit{discriminative tasks}. As we will discuss in later sections, a significant portion of NLP tasks is in fact discriminative tasks. We hypothesize that discriminative approaches perform better for discriminative tasks.

To verify the hypothesis, we propose the \textbf{universal discriminator (UD)}, which substantially improves zero-shot generalization over the previous generative state-of-the-art (SOTA)~\cite{T0-paper}, as Figure~\ref{fig:sota} shows.
The main idea is to train a single discriminator to predict whether a text sample comes from the true data distribution of natural language, similar to GANs \cite{goodfellow2014generative}. Given a set of training tasks with labeled data, we construct a dataset with positive and negative examples, where positive ones are in-distribution natural language samples and negative ones are out-of-distribution. There are two major types of discriminative tasks. The first type is tasks with multiple options, such as multi-choice question answering and news classification. We fill the options into the sentences and the ones with correct options are considered positive samples. The second type is tasks with yes/no options, which can be formulated as a binary discrimination problem itself. For example, natural language inference aims to predict whether a premise entails a hypothesis. In this case, we use a prompt to concatenate the premise $A$ and the hypothesis $B$ into a sentence ``Premise: $A$. Hypothesis: $B$.'' If entailment holds, this sample is treated as positive in-distribution samples and otherwise negative out-of-distribution ones.

For the performance of zero-shot generalization, our approach achieves new state-of-the-art on the T0 benchmark, outperforming T0 by 16.0\%, 7.8\%, and 11.5\% respectively on different scales. 
UD also achieves state-of-the-art performance on a wide range of supervised NLP tasks, using only 1/4 parameters of previous methods.
Compared with the previous generative prompt-based methods, our universal discriminator requires minimal prompting, which is simple, robust, and applicable in real-world scenarios.

In addition, we also generalize UD to a larger scope of tasks, such that UD can perform discriminative and generative tasks at the same time. Specifically, we extend UD to the encoder-decoder architecture for training on generative tasks, and restrict the model's prediction on "yes"/"no" tokens for jointly training discriminative tasks. Results prove that generalized UD maintains UD's advantages on discriminative tasks and achieves comparable results on generative tasks (See \S~\ref{sec:generalizedud}).

\section{Related Work}

\subsection{Zero-Shot Generalization Using PLMs}
Pretrained language models (PLM) can transfer knowledge from training data to downstream tasks.
Prompting methods further narrow the gap between training data and downstream tasks. \citet{PET-paper} reformulate NLP tasks into cloze filling using prompts so that PLMs can conduct zero-shot inference by generating tokens given prompted inputs. \citet{meng2022generating} use PLMs to generate class-conditioned texts with the guidance of prompts without seeing any task-specific data.
Most recently, researchers have introduced natural language prompts to unify various kinds of tasks and propose a multi-task prompted training framework to achieve great zero-shot performance even faced with unseen downstream tasks (\citet{FLAN,T0-paper,flant5}).
However, zero-shot learning has been dominated by generative approaches.

\subsection{Prompt-based and Prompt-free Methods in NLP}
Prompting is the method of reformatting NLP tasks using natural language templates to adapt to downstream tasks \cite{T5-paper,PET-paper}.
To reduce the instability and labor costs brought by prompting, researchers have tried various approaches (\citet{ptuning-paper,he2021towards}) to learn continuous prompts. 

Recently, prompt-free methods are also being explored. \citet{mahabadi2022prompt} adopts task-specific adapters to learn task descriptions implicitly for few-shot learning with PLMs. 
It has also been indicated that using null prompts without task-specific templates can achieve decent performance compared with manually-designed prompts on various tasks (\citet{logan2021cutting}).

Our work further shows that those widely used lengthy instructive prompts are not necessary for zero-shot learning. Actually, minimal prompting performs better with our discriminative formulation in the multi-task zero-shot learning setting.

\subsection{Discriminative Models in NLP}

PLMs trained with masked language modeling (MLM) \cite{devlin2018bert,liu2019roberta} can be finetuned in a discriminative manner for downstream tasks. 
ELECTRA \cite{clark2020electra} trains a discriminator to detect whether a token has been replaced. WKLM \cite{xiong2019pretrained} employs an entity-centric approach for pretraining and predicts whether an entity has been replaced.
However, finetuning for these methods is usually based on one separate CLS head per task, which is not suitable for zero-shot generalization.

Recently, prompting has been combined with token-level discriminators based on ELECTRA for few-shot learning \cite{yao2022prompt,xia2022prompting}. While these are also discriminative approaches, there are a few key differences from our approach. The biggest difference between them and us is that: we unify all discriminative tasks into one single task with minimal prompting, showing extremely good zero-shot generalization. Moreover, these methods are specific to ELECTRA-like pretraining, while our approach accepts arbitrary pretrained encoders. In our experiments, we will also make a direct comparison with these approaches to demonstrate our effectiveness.

\begin{figure*}[t]
     \centering
     \includegraphics[width=1.0\linewidth]{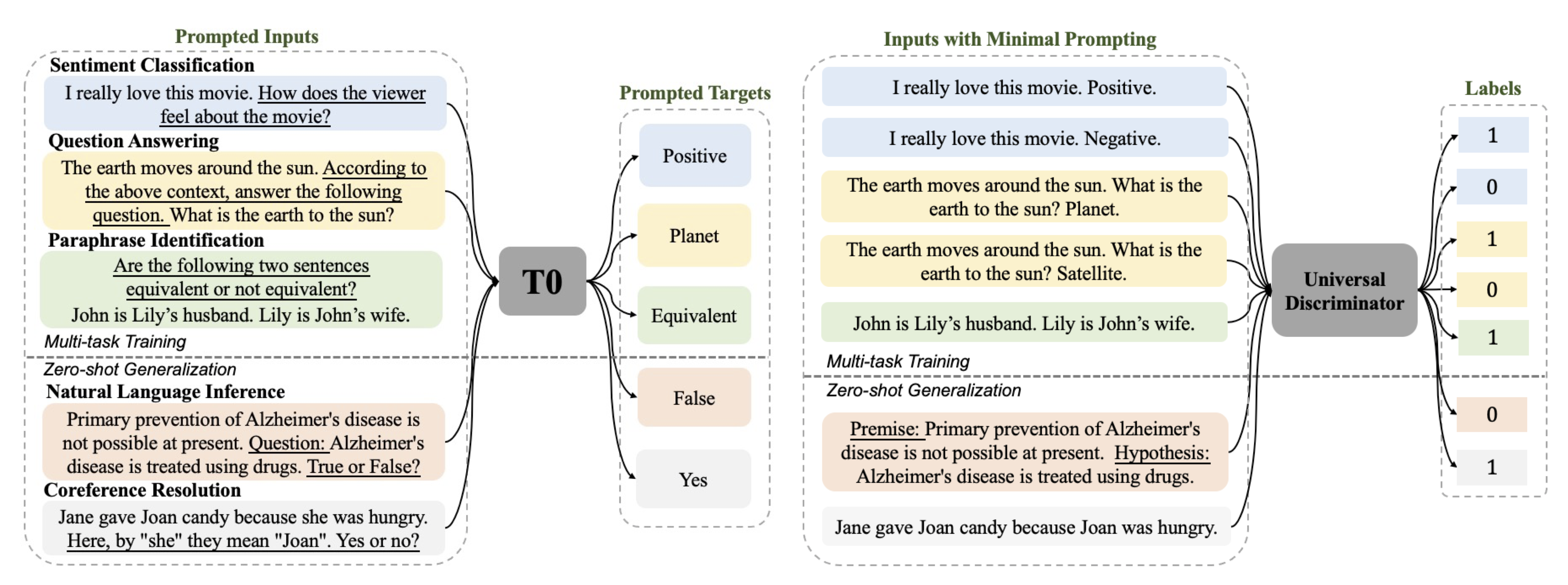}
     \caption{An overview that compares the multi-task prompted formulation of T0~\cite{T0-paper} and the formulation of our universal discriminator. The underlines mark natural language prompts. The universal discriminator uses a shared formulation of the discriminative tasks---determining whether a sample comes from the true data distribution of natural language.}
     \label{fig:overview}
     \vspace{-0.5cm}
 \end{figure*}

\section{Approach}

Previous works \citep{T0-paper,FLAN} have shown that prompted multi-task training can greatly improve zero-shot performance on unseen tasks. One intuitive reason behind the validity of this improvement is that all the NLP tasks share a common ability that allows LMs to solve unseen tasks based on the data from other training tasks. To test this idea and even enhance zero-shot generalization, a direct way is explicitly defining what this "common ability" is. Here, we define this "common ability" by designing a new general task of ``discriminating whether a text sample comes from the true data distribution of natural language''. 

We will first formulate the learning problem (\S~\ref{sec:formualtion}), and then define the concept \textit{discriminative tasks} (\S~\ref{sec:disc}), followed by describing how we transform discriminative tasks into our shared formulation.
In \S~\ref{sec:ud} and \S~\ref{sec:generalizedud}, we will study our UD, respectively on discriminative tasks and on a generalized scope of both discriminative and generative tasks.

\subsection{Multi-Task Training for Zero-Shot Generalization} \label{sec:formualtion}

Now we describe the learning problem we aim to solve in this work.
We adopt the same setting as in \citet{T0-paper}. The input to our problem is a set of training tasks with labeled data, and the goal is to train a model that generalizes to unseen test tasks. The training and test tasks are constrained to have distinct task types for the evaluation of cross-task-type generalization. A pre-trained model is jointly trained on the set of training tasks and directly evaluated on the set of test tasks in a zero-shot manner.

\subsection{Discriminative Tasks} \label{sec:disc}

We use the term ``discriminative tasks'' to refer to tasks that can be framed as selecting from a few options. 

More concretely, there are two types of discriminative tasks. The first type is tasks with multiple options, such as multi-choice question answering and news classification. The problem can be framed as selecting the right option from multiple ones, where the options are either customized for each sample (e.g., multi-choice question answering) or shared within the task (e.g., news classification). The second type is tasks with yes/no options, such as paraphrase identification and natural language inference. Given a sample of these tasks, a model is asked to predict a yes/no (or true/false) answer. 

It is important to notice that discriminative tasks constitute a significantly large portion of modern NLP research tasks. For example, all of the test tasks of the T0 benchmark~\cite{T0-paper}, SuperGLUE~\cite{wang2019superglue}, GLUE~\cite{wang2019glue}, and 85+\% tasks in BBH benchmark~\cite{bbh} are discriminative tasks.

Also note that our definition of discriminative tasks has a larger scope compared to the conventional notion of ``classification'' which usually refers to tasks with a non-customized, fixed set of labels. In contrast, discriminative tasks might have sample-customized options, e.g., multi-choice question answering and coreference resolution.

\subsection{A Universal Discriminator}
\label{sec:ud}

Given a text sample $x$, let $P(\text{true} | x)$ be the probability that $x$ is sampled from the true data distribution of natural language. We train a universal discriminator (UD), denoted as $D(x)$, to estimate the probability $P(\text{true} | x)$ for each text sample $x$. From another perspective of contrastive learning \cite{oord2018representation}, this problem can also be viewed as learning a partial order of the probability distribution. Specifically, for two text samples $x_1$ and $x_2$, if $P(\text{true} | x_1) > P(\text{true} | x_2)$, the UD is expected to predict $D(x_1) > D(x_2)$. This contrastive view is essential for tasks with multiple options, i.e., learning to select from a few options based on the partial order given by UD.

Figure~\ref{fig:overview} compares the multi-task prompted formulation of T0 and the formulation of our UD.
In the following, we will show how we use this formulation of UD to unify and solve discriminative tasks.

\subsubsection{Unifying Discriminative Tasks}
\label{sec:unifydiscriminativetasks}

We assume that for any task, the concatenation of input and the correct option follows the true data distribution of natural languages, while the concatenation of input and the other wrong options deviates much from the true data distribution.

Given this assumption, we claim that almost all discriminative tasks are equivalent to our defined task (i.e., estimating $P(\text{true} | x)$) above. Here, ``equivalent'' has bi-directional meanings: on one hand, there exists a reduction\footnote{In complexity theory, a reduction is an algorithm transforming one problem A into another problem B such that a solution for problem B could also be used to solve problem A.} from UD's task (say, task U) to any discriminative task (say, task A): given a piece of labeled training data for task A, we can generate several pieces of labeled training data for task U.

On the other hand, there exists another reduction from any discriminative task A to UD's task U: given a piece of testing data for task A, we can generate several pieces of testing data for task U such that by first predicting $D(\cdot)$ on them and then using a mapping from task U's outputs to task A's outputs, we can generate the answer for task A.

Based on the definition of discriminative tasks in \S~\ref{sec:disc}, there are two main categories, multi-choice tasks and yes/no tasks. We will discuss each category in detail as follows (also see Table \ref{tab:task_formulate_example} in appendix for specifics).

\paragraph{Multi-Choice Tasks}
For multi-choice tasks, we concatenate the text input $x_{in}$ with each choice $\{c_i\}_{i=1}^{N_c}$ to form samples. For example, for multi-choice question answering, we concatenate the given paragraph and question with each answer candidate. See Table \ref{tab:task_formulate_example} for more task formulations. During training, the concatenated samples with the correct choice are given label $1$ (true) for UD and the other incorrect ones are given label $0$ (false). During testing, similarly, we concatenate the text input 
$x_{in}$ with each choice $\{c_i\}_{i=1}^{N_c}$ 
to form several samples 
$\{(x_{in},c_i)\}_{i=1}^{N_c}$ 
and ask UD for their $D(\cdot)$ scores. We then select the sample with the maximal $D(\cdot)$ score and output its corresponding choice.

\paragraph{Tasks with Yes/No Choices}
For yes/no tasks, we directly treat the text input $x_{in}$ as a sample and assign its 0/1 label based on its yes/no label. During training, we use $x_{in}$ with its assigned 0/1 label as UD's training data. During testing, we first get the output of UD on $x_{in}$, $D(x_{in})$, and then output answer yes/no based on whether $D(x_{in})>0.5$\footnote{We note that more delicate threshold search might be possible, but we find it performs well using a constant 0.5.}. 

Empirical experiments suggest that unifying tasks with Yes/No choices in such a new way can produce better zero-shot performance than using the same method for Multi-Choice Tasks. We provide two justifications here: First, the Yes/No answer tokens here don't contain specific information and thus the model cannot benefit from concatenation. Second, the two tokens Yes/No are asymmetric in the training dataset which may result in the model uniformly assigning higher scores for one of them no matter what the task input is.

\paragraph{Minimal Prompting} A key principle we follow for task formulation is minimal prompting. From Table \ref{tab:task_formulate_example}, one can see that our prompts are minimal in the sense that they are mostly just concatenations of different elements from the raw input, discarding most of the previously instructive prompting words. This is very different from T0 \cite{T0-paper} and other generative approaches \cite{gpt3-paper,PET-paper} that add lengthy task descriptions with different wordings into the prompts.

We argue that there are two major benefits of minimal prompting. First,  previous work \cite{liu2021gpt} has shown that zero-shot and few-shot performances are very sensitive to the prompts used for inference. Minimal prompting is more robust and requires less prompt engineering efforts at test time. This is especially important for true zero-shot real-world applications as there is no data available for choosing the right prompt. Second, as we will show in our experiments, UD performs much better with minimal prompts than lengthy descriptive prompts, while generative approaches do not work well with minimal prompts. This is also consistent with our motivation that all the NLP tasks share a common ability: ``discriminating whether a text sample comes from the true data distribution'' and UD is attempting to learn ``what kind of concatenation between input and option makes it look like the true language?'', which does not rely much on the descriptions for each task. On the other hand, T0 attempts to generate the answer directly basing on all the information it gets, so prompts provide an extra source of information and are helpful. See \S~\ref{sec:minimal_prompts} for our ablation study on minimal prompts.

Note that it is also important to use minimal prompts to resolve ambiguity in some cases. For example, consider the natural language inference (NLI) task that predicts whether a premise $A$ entails a hypothesis $B$. Simply concatenating $A$ and $B$ is ambiguous, because the model cannot tell which is the premise. The model also is not aware that this is an NLI task. To resolve this kind of ambiguity, we use a minimal prompt ``Premise: A. Hypothesis: B.'' instead, as shown in Table \ref{tab:task_formulate_example}.

\subsubsection{Architecture}

UD can use any pre-trained encoder model as the backbone. In this work, we experiment with the T5 encoder and DeBERTa \cite{debertav3}. Since T5 is an encoder-decoder model, we only use the encoder part. For the T5 backbone, we perform mean pooling over the last-layer encoder features, followed by a dropout layer and a linear layer to predict a scalar logit. For the DeBERTa backbone, we use the last-layer feature of the first token, followed by a two-layer perceptron with dropout to also output a scalar logit. We train UD with the binary cross entropy loss.

\subsection{A Generalized Universal Discriminator}
\label{sec:generalizedud}

To further study how the discriminative approaches work in combination with generative tasks, we also propose to experiment with a generalized version of UD (denoted as generalized UD).

Different from the previous UD that only uses an encoder as the backbone model, the generalized UD employs an encoder-decoder architecture. In the following, we experiment with the T5 model.
Generalized UD takes both discriminative and generative tasks into consideration, and is jointly trained over both types of tasks at the same time.

For discriminative tasks, they are reformulated into binary classification tasks through minimal prompting, as is described in  \S~\ref{sec:unifydiscriminativetasks}. Specifically, it takes the minimal prompted texts into the encoder and uses the decoder to predict over \{``Yes'', ``No''\}.
In such cases, generalized UD is optimized with the binary cross-entropy loss.
For generative tasks, they take the form of ``input-and-target'' pairs. Generalized UD is fed with the textual inputs, and generates the targets through decoding.
For generative tasks, generalized UD is trained to optimize the cross-entropy loss.

\begin{table*}[t]
\setlength{\tabcolsep}{1.5mm}
\centering

\subtable[On 11 discriminative test tasks following the T0 benchmark.]{
\resizebox{\textwidth}{!}{%
    \begin{tabular}{l|l|c|ccccc|ccc|cc|c|c}
    \toprule[1pt]
    \multirow{2}{*}{Base Model} &
    \multirow{2}{*}{Method} &
    \multirow{2}{*}{\#Params} & 
    \multicolumn{5}{|c|}{\textbf{Natural Language Inference}} & \multicolumn{3}{|c|}{\textbf{Sentence Completion}} & \multicolumn{2}{c|}{\textbf{Coreference}} & \multicolumn{1}{c|}{\textbf{WSD}} 
    & \multirow{2}{*}{Avg.}\\
    & & & RTE & CB & ANLI1 & ANLI2 & ANLI3 & COPA & Hella. & Story. & WSC & Wino. & WiC &  \\
    \midrule[1pt]
    Decoder-only & GPT-3 & 175B 
        &63.5 &46.4
        &34.6 &	35.4&	34.5&	91.0&	78.9&	83.2&	65.4&	70.2&	- & -\\
    Decoder-only & GLaM & 137B 
        & 56.3	& 39.3	& 39.7	& 35.5	& 34.1	& 90.0	& 76.7	& 81.1	& 82.1	& 71.3	& 50.6 & 59.7\\
    MoE Decoder-only & GLaM & 64B 
        & 66.8	& 33.9	& 40.9	& 38.2	& 40.9	& 90.0	& 77.1	& 82.5	& 83.5	& 73.4	& 50.5 & 61.6\\
    Decoder-only & PaLM & 540B 
        & 72.9	& 51.8	& 48.0	& 44.2	& 45.7	& 93.0	& 83.4	& 84.6	& 89.1	& 81.1	& 59.1 & 68.5\\
    Decoder-only & FLAN & 137B 
        & 78.3	& 64.1	& 47.7	& 43.9	& 47.0	& 90.6	& 56.4	& 92.2	& 80.8	& 67.3 & - & -\\
    \midrule[1pt]
    \multirow{3}*{\shortstack{ELECTRA}}
    & PE-CLS & 335M
        & 60.2	& 57.4	& 34.1	& 34.4	& 36.4	& 92.7	& 44.1	& 96.0	& 62.8	& 56.3	& 50.7	& 56.8
        \\
    & PE-PROB & 335M
        & 54.0	& 49.2	& 32.3	& 33.3	& 33.5	& 81.9	& 36.7	& 89.5	& 64.3	& 50.7	& 50.9	& 52.4 \\
    & PE-REP & 335M
        & 69.0	& 61.3	& 36.1	& 35.0	& 39.4	& 91.2	& 47.0	& 96.8	& 70.0	& 56.2	& 51.1	& 58.5
        \\
    \midrule
    \multirow{1}*{\shortstack{DeBERTaV3}}
    & \multirow{1}*{{UD (ours)}} & 304M
        & \multirow{1}*{71.1}
        & \multirow{1}*{76.8}
        & \multirow{1}*{43.8}
        & \multirow{1}*{41.3}
        & \multirow{1}*{45.7}
        & \multirow{1}*{96.0}
        & \multirow{1}*{60.7}
        & \multirow{1}*{97.4}
        & \multirow{1}*{66.4}
        & \multirow{1}*{83.6}
        & \multirow{1}*{53.3}
        & \multirow{1}*{66.9}
    \\
    \midrule[1pt]
    \multirow{2}*{\shortstack{T5-Large}}
    & \multirow{1}*{T0 $\star$} & 800M
        & 75.1	& 55.5	& 32.9	& 32.3	& 33.7	& 84.6	& 28.2	& 94.0	& 63.0	& 54.6	& 51.2	& 55.0 \\

    & {UD (ours)} & 400M
        & \textbf{83.8}
        & \textbf{80.4}
        & \textbf{36.8}
        & \textbf{34.2}
        & \textbf{42.2}
        & \textbf{90.0}
        & \textbf{56.1}
        & \textbf{96.4}
        & \textbf{68.3}
        & \textbf{62.9}
        & \textbf{54.6}	
        & \textbf{64.1} \\
    \midrule[1pt]
    \multirow{3}*{\shortstack{T5-XL}}
    & \multirow{1}*{T0 $\dagger$} & 3B
        & 64.6 
        & 45.4
        & 33.8
        & 33.1
        & 33.3
        & 72.4
        & 27.3
        & 84.0
        & 65.1
        & 51.0
        & 50.7
        & 51.0 \\

    & \multirow{1}*{T0 $\star$} & 3B
    & \textbf{79.7}	& 68.9	& \textbf{43.1}	& \textbf{38.5}	& 42.3	& \textbf{94.1}	& 31.5	& 97.5	& 68.8	& 61.3	& \textbf{54.1}	& 61.8\\

    & {UD (ours)} & 1.5B
        & 78.7
        & \textbf{73.2}
        & 41.2
        & 36.3
        & \textbf{45.4}
        & 94.0
        & \textbf{70.1}
        & \textbf{97.9}
        & \textbf{72.1}
        & \textbf{70.6}
        & 53.0	
        & \textbf{66.6} \\
    \midrule[1pt]
    \multirow{4}*{\shortstack{T5-XXL}}
    & \multirow{1}*{T0 $\dagger$} & 11B
        & 80.8
        & 70.1
        & 43.6
        & 38.7
        & 41.3
        & 90.0
        & 33.6
        & 92.4
        & 61.5
        & 59.9
        & 56.6
        & 60.8 \\

    & \multirow{1}*{T0 $\star$} & 11B
    & \textbf{85.8}	& 73.3	& 47.3	& 42.0	& 46.1	& 94.4	& 31.5	& 98.4	& 62.8	& 72.8	& 56.0	& 64.6 \\

    & {UD (ours)} & 5.5B
    & 80.5	& 87.5	& 49.0	& 42.9 & 	48.8	& 95.0	& 77.4	& \textbf{98.6}	& 73.1	& 82.2	& 57.1	& 72.0 \\

    & {UD+ (ours)} & 5.5B
    & 82.0	& \textbf{89.3}	& \textbf{53.4} & \textbf{48.1} & \textbf{51.0} & \textbf{96.0} & \textbf{78.9} & 96.7	& \textbf{75.0}	& \textbf{86.4}	& \textbf{58.5}	& \textbf{74.1} \\
    \bottomrule[1pt]
\end{tabular}
}
\label{tab:maintable:top}
}

\subtable[On 13 discriminative BigBench tasks following the T0 benchmark]{
\resizebox{0.7\textwidth}{!}{%
    \begin{tabular}{l|cc|cc|ccc|}
    \toprule[1pt]
    \multirow{1}{*}{Model} 
        & \multirow{1}{*}{\shortstack{T0-Large}}
        & \multirow{1}{*}{\shortstack{UD-large}}
        & \multirow{1}{*}{\shortstack{T0-XL}}
        & \multirow{1}{*}{\shortstack{UD-XL}}
        & \multirow{1}{*}{\shortstack{T0-XXL}}
        & \multirow{1}{*}{\shortstack{UD-XXL}}
        & \multirow{1}{*}{\shortstack{UD+-XXL}}\\
    \midrule[1pt]
    BigBench (Avg.) & 39.6 & \textbf{43.5} & 44.8 & \textbf{48.9} & 47.4 & 55.5 & \textbf{58.7} \\
    \bottomrule[1pt]
    \end{tabular}%
    }
\label{tab:maintable:middle}
}

\subtable[On 22 discriminative BBH tasks]{
\resizebox{\textwidth}{!}{%
    \begin{tabular}{l|ccc|ccc|cccc|}
    \toprule[1pt]
    \multirow{1}{*}{Model} 
        & \multirow{1}{*}{\shortstack{T0-Large}}
        & \multirow{1}{*}{\shortstack{Flan-T5-Large}}
        & \multirow{1}{*}{\shortstack{UD-Large}}
        & \multirow{1}{*}{\shortstack{T0-XL}}
        & \multirow{1}{*}{\shortstack{Flan-T5-XL}}
        & \multirow{1}{*}{\shortstack{UD-XL}}
        & \multirow{1}{*}{\shortstack{T0-XXL}}
        & \multirow{1}{*}{\shortstack{Flan-T5-XXL}}
        & \multirow{1}{*}{\shortstack{UD-XXL}}
        & \multirow{1}{*}{\shortstack{UD+-XXL}}\\
    \midrule[1pt]
    BBH (Avg.) & 38.9 & 39.5 & \textbf{44.2} & 40.4 & 44.6 & \textbf{47.3} & 45.0 & 49.4 & 51.3 & \textbf{56.7} \\
    \bottomrule[1pt]
    \end{tabular}%
    }
\label{tab:maintable:bottom}
}
\caption{
Zero-shot performance of our UD and baselines.
Results in the first block are reported by previous work, respectively from GPT-3~\cite{gpt3-paper}, GLaM~\cite{glam}, PaLM~\cite{palm}, and FLAN~\cite{FLAN}.
Note that we provide these reported results for reference, and do not compare directly. Some of the reported tasks are evaluated on the test split, while we follow the better baseline method T0 to report on validation splits.
Results with $\dagger$ are reported by~\citeauthor{T0-paper}, and results with $\star$ are reproduced in our framework. We reproduced the three variants of prompting ELECTRA~\cite{xia2022prompting} under our setting, denoted as ``PE-CLS'', ``PE-PROB'', ``PE-REP''.
Results for Flan-T5-Large/Xl/XXL~\citep{flant5} are reproduced by testing zero-shot performance on their released checkpoints.
In the same group, T0 and Flan-T5 has 2x model parameters compared to UD. For abbreviation, we denote UD based on T5-XX as ``UD-XX'', e.g., UD-XL refers to UD based on the T5-XL model.
}
\label{tab:maintable}
\vspace{-0.7cm}
\end{table*}

\begin{table}[htbp]
\setlength{\tabcolsep}{1.5mm}
  \centering
\resizebox{0.35\textwidth}{!}{
    \begin{tabular}{lcc}
    \toprule
    \textbf{Dataset} & \textbf{SOTA} & \textbf{UD+-XXL} \\
    \midrule
    QQP     & \textbf{90.60}  & 90.44 \\
    DREAM     & 91.80  & \textbf{94.95} \\
    QuAIL   & 87.20  & \textbf{88.13} \\
    IMDB    & 97.30  & \textbf{97.44}  \\
    AgNews   & \textbf{95.58}  & 95.56  \\
    OBQA   & 87.20  & \textbf{89.20} \\
    STSB     & 92.30  & \textbf{92.90} \\
    CSQA    & \textbf{84.90}  & 84.68  \\
    SST-2     & 97.30  & \textbf{97.48} \\
    QNLI    & 96.50  & \textbf{96.56} \\
    AbductiveNLI &  89.80  & \textbf{93.20} \\
    VitaminC   & 91.10  & \textbf{92.62} \\
    MNLI  &  \textbf{92.10}  & 92.03  \\
    MCScript &  97.30  & \textbf{98.03} \\
    MCScript 2.0 &  97.90  & \textbf{98.01} \\
    AdversarialNLI (r3) &53.50  & \textbf{67.83 } \\
    COLA   & \textbf{71.50}  & 71.42  \\
    \midrule
    Avg.   & 89.05  & \textbf{90.62} \\
    \bottomrule
    \end{tabular}%
}
  \caption{Results on fully-supervised tasks for UD, which is based on the encoder of T5-xxl. Previous sota model \citep{ul2} has 4x model parameters compared to UD. }
  \label{tab:finetune}%
\vspace{-0.7cm}
\end{table}%

\section{Experiments}

\begin{table*}[t]
\setlength{\tabcolsep}{1.5mm}
\centering
\small
\resizebox{\textwidth}{!}{%
    \begin{tabular}{l|ccccc|ccc|cc|c|c}
        \toprule[1pt]
        & \multicolumn{5}{c|}{\textbf{Natural Language Inference}} & \multicolumn{3}{|c|}{\textbf{Sentence Completion}} & \multicolumn{2}{c|}{\textbf{Coreference}} & \multicolumn{1}{c|}{\textbf{WSD}} & \multirow{2}{*}{Avg.} \\
    & RTE & CB & ANLI1 & ANLI2 & ANLI3 & COPA & Hella. & Story. & WSC & Wino. & WiC &  \\
    \midrule[1pt]
    UD (Minimal)     & \textbf{83.8}
        & \textbf{80.4}
        & 36.8
        & \textbf{34.2}
        & \textbf{42.2}
        & \textbf{90.0}
        & \textbf{56.1}
        & \textbf{96.4}
        & \textbf{68.3}
        & \textbf{62.9}
        & \textbf{54.6}	
        & \textbf{64.1} \\
    UD (Instructive)    & 72.2 
        & 64.5 
        & \textbf{37.0} 
        & 33.4 
        & 39.7 
        & 85.3 
        & 45.2 
        & 96.0 
        & 65.4 
        & 53.9 
        & 50.9 
        & 58.5\\
    \midrule
    T0 (Minimal) & 61.6  & \textbf{57.8}  & 30.6  & 30.3  & 33.4  & 67.2  & \textbf{33.8}  & 66.6  & 60.9  & 52.8  & \textbf{51.7}  & 49.7  \\
    T0 (Instructive) & \textbf{75.1}	& 55.5	& \textbf{32.9}	& \textbf{32.3}	& \textbf{33.7}	& \textbf{84.6}	& 28.2	& \textbf{94.0}	& \textbf{63.0}	& \textbf{54.6}	& 51.2	& \textbf{55.0} \\

    \bottomrule[1pt]
    \end{tabular}}
    \caption{Zero-shot performance for UD and T0 respectively with instructive and minimal prompts. Instructive prompts are lengthy descriptions of tasks \citep{T0-paper}, while minimal prompts use a simple concatenation of input data.}
\label{tab:promptablatiion}
\vspace{-0.5cm}
\end{table*}

\subsection{Experimental Setup}\label{sec:setup}

We performed extensive experiments to validate the performance of the zero-shot generalization of our UD. We follow the same zero-shot setting as T0~\citep{T0-paper} by training on multi-task datasets and evaluating a held-out set of tasks that are never seen during training. 

\paragraph{Datasets}
The original T0 training set consists of 38 tasks of 8 different types.
There are in total 21/38 discriminative training tasks, with which we train the UD.
The evaluation set covers four types of tasks, including natural language inference (RTE~\citep{2005_RTE}, CB~\citep{de2019_CB}, ANLI/R1-R3~\citep{NieWDBWK20_ANLI}), coreference resolution (WSC~\citep{WSC2012}, Winogrande~\citep{SakaguchiBBC20_winogrande}), sentence completion (COPA~\citep{COPA2011}, StoryCloze~\citep{story_cloze}, Hellaswag~\citep{ZellersHBFC19_hellaswag}), and word sense disambiguation (WiC~\citep{wic-paper}).
Following T0, we use accuracy on the validation split as the evaluation metric.
For prompt-based baselines, we report the average accuracy over multiple prompts for each test task.
Besides, we also evaluate zero-shot performance on 13 BigBench~\cite{bigbench} tasks, which are also adopted by T0~\cite{T0-paper}, %
and 22 BBH tasks~\cite{bbh}, which are adopted by Flan-T5~\cite{flant5}.

\paragraph{Baselines}
We primarily compare our method with T0~\citep{T0-paper}, which is a generative approach.
Another baseline is prompting ELECTRA~\cite{xia2022prompting} which is a recent work on discriminative modeling.
Since it was proposed in a different setting (i.e., a  few-shot setting or direct zero-shot inference without any finetuning), we reproduced their method under our multitask zero-shot setting for comparison.

For a fair comparison, we follow T0 to use the T5-V1.1-LM-Adapted~\citep{T5-paper} as the backbone model, and we experimented with three different scales, respectively 800M, 3B, and 11B. 
For UD, it only makes use of the encoder of T5-v1.1 and additionally replaces the output layer with a classification head.

In addition, we provide reported zero-shot results of several large language models (with hundreds of billions of parameters) for reference, including GPT-3~\cite{gpt3-paper}, GLaM~\cite{glam}, PaLM~\cite{palm}, and FLAN~\cite{FLAN}. We also reproduce zero-shot results of a recent work Flan-T5~\cite{flant5} by evaluating their released checkpoints on BBH tasks\footnote{T0 test sets are included in Flan-T5's  training data sets, so we can't test its zero-shot performance on those data sets.}. Note that Flan-T5's training data sets are much broader than ours, so results for Flan-T5 here are only for reference but not a fair comparison.

\paragraph{Training}
During training, we truncate the input sequence to 256 tokens and use a batch size of 256. For optimization, we use the Adam optimizer with a fixed learning rate of 1e-5 and a dropout rate of 0.1. Each experiment is trained with 10, 8, and 5 epochs respectively for 800M, 3B, and 11B models.

\subsection{Main Results on Zero-Shot Tasks}

\paragraph{UD Zero-Shot Results}
The main results are presented in Table~\ref{tab:maintable}.
We compare methods of similar scales. 
Results in Table~\hyperref[tab:maintable:top]{1(a)} show that our UD substantially outperforms the T0 baseline on average by a large margin of around 9, 5, and 7 points respectively at Large, XL, and XXL scales.
Comparing the results of UD-T5-Large, UD-DeBERTaV3, and prompting ELECTRA, both variants of UD also substantially outperform prompting ELECTRA by more than 6 points.
On BIG-Bench datasets, results in Table~\hyperref[tab:maintable:middle]{1(b)} show that our UD outperforms the T0 baseline by a margin of around 4-8 points.
Besides T0 benchmark, we also test UD on BBH datasets, which are very different from T0 training sets, results in Table~\hyperref[tab:maintable:bottom]{1(c)} show that our UD constantly outperforms T0 and Flan-T5 by a margin of around 2-5 points, even though our UD is only trained on a small fraction of Flan-T5's training sets.
Overall, these results demonstrate the advantages of UD at every scale, and a broad range of tasks compared with baselines.

Another interesting finding is that the advantages of UD significantly increase along with scaling.
When scaling from Large-scale to XL-scale (i.e., around 3.75x of the parameters), the average performance improves by around 2 points. However, when scaling from XL-scale to XXL-scale (i.e., 3.6x of the parameters), the improvements of average zero-shot performance enlarge to 8 points.
Based on the observation, we hypothesize that UD can achieve even better performance of zero-shot generalization if further scaling to an even larger models, which we leave to future work.

To further boost the zero-shot performance, we also train a new variant of UD at 11B scale by scaling to more training tasks, including the discriminative English tasks used in \citet{1600tasks}, and the discriminative English tasks used in \citet{ul2}. The new model is denoted as UD+.
UD+ achieves the highest average accuracy among all the zero-shot evaluation tests.

\paragraph{Generalized UD Zero-Shot Results}
\begin{table*}[!htp]
\setlength{\tabcolsep}{1.5mm}
\centering
\subtable[On 11 discriminative test tasks following the T0 benchmark.]{
\resizebox{\textwidth}{!}{%
    \begin{tabular}{l|ccccc|ccc|cc|c|c}
        \toprule[1pt]
        \multirow{2}*{Method}
        & \multicolumn{5}{c|}{\textbf{Natural Language Inference}} & \multicolumn{3}{c|}{\textbf{Sentence Completion}} & \multicolumn{2}{c|}{\textbf{Coreference}} & \multicolumn{1}{c|}{\textbf{WSD}} & \multirow{2}{*}{Avg.} \\
    & RTE & CB & ANLI1 & ANLI2 & ANLI3 & COPA & Hella. & Story. & WSC & Wino. & WiC &  \\
    \midrule[1pt]
    T0-XL %
        & \textbf{79.7}	& 68.9	& 43.1	& 38.5	& 42.3	& \textbf{94.1}	& 31.5	& \textbf{97.5}	& \textbf{68.8}	& 61.3	& \textbf{54.1}	& 61.8\\
    GenUD-XL %
        & 71.5	& \textbf{80.4}	& \textbf{43.1}	& \textbf{39.5}	& \textbf{42.6}	& 94.0	& \textbf{55.8}	& 96.7	& 63.5	& \textbf{75.5}	& 52.8	& \textbf{65.0}\\
    \bottomrule[1pt]
    \end{tabular}%
}
\label{tab:genud:top}
}
\subtable[On 13 discriminative Big-Bench tasks following the T0 benchmark.]{
\resizebox{\textwidth}{!}{%
    \begin{tabular}{l|ccccccccccccc|c}
    \toprule[1pt]
    \multirow{2}{*}{Model} 
        & \multirow{2}{*}{\shortstack{code \\ desc.}}
        & \multirow{2}{*}{\shortstack{conce\\-ptual}}
        & \multirow{2}{*}{\shortstack{known\\unknowns}}
        & \multirow{2}{*}{\shortstack{logic \\ grid}}
        & \multirow{2}{*}{\shortstack{logic \\ deduction}}
        & \multirow{2}{*}{\shortstack{miscon\\-ceptions}}
        & \multirow{2}{*}{\shortstack{novel\\concepts}}
        & \multirow{2}{*}{\shortstack{strate\\-gyqa}}
        & \multirow{2}{*}{\shortstack{wino\\-why}}
        & \multirow{2}{*}{\shortstack{syllo\\-gisms}}
        & \multirow{2}{*}{\shortstack{movie\\dialog}}
        & \multirow{2}{*}{\shortstack{lang\\-uage\_id}}
        & \multirow{2}{*}{\shortstack{vita\\-minc}} 
        & \multirow{2}{*}{Avg.} \\
    &&&&&&&&&&&&&&\\
    \midrule
    T0-XL & 23.4 & 48.1 & 64.6 & \textbf{42.5} & 50.1 & \textbf{52.7} & 25.0    & 53.1 & 45.4 & 50.2 & 47.7 & \textbf{19.0} & 60.0 & 44.8 \\
    GenUD-XL & \textbf{60.0} & \textbf{64.1} & \textbf{69.6} & 38.2 & \textbf{52.8}  & 48.9 & \textbf{44.1} & \textbf{57.1} & \textbf{46.5} & \textbf{50.4} & \textbf{50.9} & 15.5 & \textbf{66.8} & \textbf{48.9} \\
    \bottomrule[1pt]
    \end{tabular}%
\label{tab:genud:mid}
}
}

\subtable[On 15 generative tasks from Big-Bench]{
\resizebox{\textwidth}{!}{%
    \begin{tabular}{l|ccccccccccccccc|c}
    \toprule[1pt]
    \multirow{3}{*}{Model}
        & \multirow{3}{*}{\shortstack{auto \\ debugging}}
        & \multirow{3}{*}{\shortstack{simple \\ arith \\ -metic}}
        & \multirow{3}{*}{\shortstack{repeat\\copy \\ logic}}
        & \multirow{3}{*}{\shortstack{sufficient \\ information}}
        & \multirow{3}{*}{\shortstack{simple \\ text \\ editing}}
        & \multirow{3}{*}{\shortstack{scientific \\ press \\ release}}
        & \multirow{3}{*}{\shortstack{code\\ names}}     
        & \multirow{3}{*}{\shortstack{emoji\\movies}}
        & \multirow{3}{*}{\shortstack{penguins\\in a \\ table}}
        & \multirow{3}{*}{\shortstack{few \\ shot\\nlg}}
        & \multirow{3}{*}{\shortstack{operators}}
        & \multirow{3}{*}{\shortstack{tense}}
        & \multirow{3}{*}{\shortstack{geometric\\shapes}}
        & \multirow{3}{*}{\shortstack{chinese \\ remainder\\ theorem}}
        & \multirow{3}{*}{\shortstack{temporal\\sequences}}
        & \multirow{3}{*}{\shortstack{Avg.}}\\
    &&&&&&&&&&&&&&&&\\[1em]
    \midrule
    T0-XL & 11.2 & 6.7 & \textbf{25.8} & 33.8 & 7.5 & \textbf{6.7} & \textbf{44.8} & \textbf{8.7} & \textbf{11.4} & 17.4 & \textbf{10.5} & 80.7 & 0.0 & 0.0 & 14.0 & \textbf{18.6}\\
    GenUD-XL & \textbf{15.5} & 6.7 & 8.2 & \textbf{34.4} & \textbf{12.6} & 6.4 & 25.1 & 0.0 & 8.1 & \textbf{20.5} & 3.7 & \textbf{80.9} & 0.0 & 0.0 & \textbf{33.5}  & 17.0\\
    \bottomrule[1pt]
    \end{tabular}%
}
}

\caption{Zero-shot performance for generalized UD and T0 on discriminative and generative tasks. 
We select the top 15 uncommon generative tasks from BigBench basing on ascending order of data size. (We assume that datasets with smaller sizes are less common, and more suitable for zero-shot tests.) The metrics are respectively accuracy for discriminative tasks and ROUGE1 for generative tasks. ``GenUD'' denotes our generalized UD method.}
\label{tab:genud}
\end{table*}

The zero-shot results of generalized UD on 11 T0 discriminative test tasks and on 13 Big-Bench tasks are respectively reported in Table~\hyperref[tab:genud:top]{7(a)} and Table~\hyperref[tab:genud:mid]{7(b)} 
We also select the top 15 uncommon generative tasks from BigBench basing on ascending order of data size, results are in Table~\hyperref[tab:genud:bottom]{7(c)}. We assume that tasks with smaller data sizes are less common and more likely to be unrelated to our training data and more suitable for zero-shot tests.

Analyses are as follows.
First, comparing the results of generalized UD and T0, generalized UD still holds significant improvements on discriminative tasks.
Second, comparing generalized UD with our previous UD (in Table~\ref{tab:maintable}), we observe there is a slight decrease in average performance, proving that adding generative tasks into training could have impacted a little bit, in trade for capability for handling generative tasks.
Third, on 15 generative tasks, both generalized UD and T0 show comparable results.

\subsection{SOTA Results on Finetuned Tasks}
\label{sec:ud_finetune}

To explore how UD performs on fully-supervised tasks, we finetuned UD for a wide range of downstream tasks and reported their results in Table \ref{tab:finetune}.
For each finetuning experiment, the maximum training epoch is set to be 10.
We search a hyper-parameter space with learning rate in \{2e-5, 1e-5, 5e-6\}, batch size in \{32, 64, 128\}.
We select the best checkpoint using a validation set with early stopping.

From results in Table \ref{tab:finetune}, we find that UD can achieve remarkable performance on most of the downstream tasks. 
We achieve state-of-the-art performance on 12 out of the 17 tasks we evaluated. The results also show that more challenging tasks (tasks that require more knowledge) will benefit more from the multi-task training period, especially some QA tasks.

\subsection{Ablation Study}

We have also conducted ablation studies to further explore how several factors affect the performance of zero-shot generalization. Please see appendix for further ablation studies on UD with different base models (\S~\ref{sec:base_models})

\subsubsection{Instructive Prompts vs Minimal Prompts}
\label{sec:minimal_prompts}

UD employs minimal prompts that use simple concatenation, while previous approaches rely on lengthy instructive prompts to provide more detailed instructions \cite{T0-paper,FLAN,gpt3-paper}. 
Statistically, we count the average number of prompt words (excluding raw input) for both minimal and instructive prompts, and statistics are respectively $0.4$ versus $>10$.
We compare these two types of prompts in the following experiment.
We adopt the instructive prompts from T0 and apply them on UD without changing the discriminator formulation. To construct minimal prompts for T0, we remove all the instructive words similar to UD.

Results are shown in Table~\ref{tab:promptablatiion}. We observe that minimal prompts yield better performance for UD than instructive prompts. In contrast, for T0, instructive prompts perform much better than minimal prompts. These results are consistent with our motivation that UD tends to unify the tasks better with a shared discrimination formulation. As a result, task-specific instructions are not necessary and might hurt generalization performance. Generative approaches, on the other hand, rely on instructive prompts to better distinguish different tasks and generate specific answers directly.

\subsection{How Well UD Generalizes to a Broader Domain?} \label{sec:generalize}
\begin{table}
\centering
\setlength{\tabcolsep}{3.0mm}
\resizebox{0.5\textwidth}{!}{%
\begin{tabular}{l|c}
    \toprule[1pt]
    Setting & Accuracy \\
    \midrule[1pt]
    True Data vs Manually-Generated Data & 80.0 \\
    True Data vs Model-Generated Data & 74.4 \\
    \bottomrule[1pt]
    \end{tabular}%
    }
    \caption{
    The accuracy of UD discriminating real data and generated data. We feed UD with a real sample $x$ from the real-world data distribution, and a sample $x'$ from manual generation or model-based generation. 
    If UD assigns higher score to $x$ than $x'$ (i.e., $D(x)>D(x')$), it is considered an accurate prediction.
    }
  \label{tab:explain}%
\end{table}%

Our discrimination problem formulation is in fact more general than solving supervised labeled tasks and can be applied to a broader domain of natural language. We conduct the following experiment to see how UD generalizes.

To test whether a model discriminates against the true data distribution, a straightforward way of verification is to compare the probability of real data with that of some generated, fake data. This form of verification is not specific to any downstream task and can be viewed as generalizing to a broader domain. Formally, given a text sample $x$, let $D(x)$ be the output of UD, which estimates the probability that $x$ is sampled from the true data distribution, i.e., $P(\text{true} | x)$. Given a true data sample $x$ and a generated data sample $x'$, we expect a well-trained UD to predict $D(x) > D(x')$.

Specifically, we randomly select 2,600 real data samples $x$ from the validation set of the T0 training data and generate the data $x’$ in two different ways: model-based generation and manual generation.

For a model-based generation, we utilize the T0-Large model with a paraphrase prefix ``Paraphrase the sentence:'' to generate data $x'$. It is expected that the generated samples $x'$ are similar to true samples $x$ to some extent but demonstrate some flaws that are unique to generated data. For a manual generation, we manually create some conflict or contradiction in the real sample $x$. Specifically, we manually attach wrong answers to the original data and obtain $x’$ , which is similar to what we have done in constructing negative samples in our main framework. 

We then use our \method based on T5-Encoder Large to compute the probability $D(x)$ and $D(x')$ for both real and generated data. As displayed in Table~\ref{tab:explain}, we find that the \method assigns a higher score for $x$ than $x'$ $80\%$ of the time for manually-generated data. When tested with model-generated data, UD assigns a high probability for real data in $74\%$ of the cases.
This is probably because manually generated data are more paradoxical and logically incoherent and thus are easier for UD to discriminate. Overall, these results demonstrate that the discrimination ability of UD is not limited to the downstream tasks on which it was trained, but is also generalizable to a broader domain of text data. This indicates a possibility of extending UD to other scenarios such as model pretraining and generation tasks.

\section{Conclusions}
Universal Discriminator is a discriminating model for predicting whether a sample comes from the true data distribution, which is a new formulation for all discriminative NLP tasks. Experiments show that UD sets the new state-of-the-art for zero-shot generalization on many benchmarks. UD is high-performing with minimal prompting, and thus is more robust and applicable in practice. A generalized UD can also solve generative tasks at the same time which keeps UD's advantage on discriminative tasks and has comparable performance on generative tasks.

\section{Limitation}
Even though our generalized UD can get comparable performance on some generative tasks, generalized UD may not handle certain complex generation tasks very well (e.g., summarization) 
We leave expanding UD to solve a broader range of generative tasks and achieve greater performance advantage as our future work.

\bibliography{anthology}

\begin{thebibliography}{41}
\expandafter\ifx\csname natexlab\endcsname\relax\def\natexlab#1{#1}\fi

\bibitem[{Artetxe et~al.(2021)Artetxe, Bhosale, Goyal, Mihaylov, Ott, Shleifer,
  Lin, Du, Iyer, Pasunuru et~al.}]{artetxe2021efficient}
Mikel Artetxe, Shruti Bhosale, Naman Goyal, Todor Mihaylov, Myle Ott, Sam
  Shleifer, Xi~Victoria Lin, Jingfei Du, Srinivasan Iyer, Ramakanth Pasunuru,
  et~al. 2021.
\newblock Efficient large scale language modeling with mixtures of experts.
\newblock \emph{arXiv preprint arXiv:2112.10684}.

\bibitem[{Brown et~al.(2020)Brown, Mann, Ryder, Subbiah, Kaplan, Dhariwal,
  Neelakantan, Shyam, Sastry, Askell, Agarwal, Herbert{-}Voss, Krueger,
  Henighan, Child, Ramesh, Ziegler, Wu, Winter, Hesse, Chen, Sigler, Litwin,
  Gray, Chess, Clark, Berner, McCandlish, Radford, Sutskever, and
  Amodei}]{gpt3-paper}
Tom~B. Brown, Benjamin Mann, Nick Ryder, Melanie Subbiah, Jared Kaplan,
  Prafulla Dhariwal, Arvind Neelakantan, Pranav Shyam, Girish Sastry, Amanda
  Askell, Sandhini Agarwal, Ariel Herbert{-}Voss, Gretchen Krueger, Tom
  Henighan, Rewon Child, Aditya Ramesh, Daniel~M. Ziegler, Jeffrey Wu, Clemens
  Winter, Christopher Hesse, Mark Chen, Eric Sigler, Mateusz Litwin, Scott
  Gray, Benjamin Chess, Jack Clark, Christopher Berner, Sam McCandlish, Alec
  Radford, Ilya Sutskever, and Dario Amodei. 2020.
\newblock Language models are few-shot learners.
\newblock \emph{CoRR}, abs/2005.14165.

\bibitem[{Candela et~al.(2006)Candela, Dagan, Magnini, and
  d'Alch{\'{e}}{-}Buc}]{2005_RTE}
Joaquin~Qui{\~{n}}onero Candela, Ido Dagan, Bernardo Magnini, and Florence
  d'Alch{\'{e}}{-}Buc, editors. 2006.
\newblock \emph{Machine Learning Challenges, Evaluating Predictive Uncertainty,
  Visual Object Classification and Recognizing Textual Entailment, First
  {PASCAL} Machine Learning Challenges Workshop, {MLCW} 2005, Southampton, UK,
  April 11-13, 2005, Revised Selected Papers}, volume 3944 of \emph{Lecture
  Notes in Computer Science}. Springer.

\bibitem[{Chowdhery et~al.(2022)Chowdhery, Narang, Devlin, Bosma, Mishra,
  Roberts, Barham, Chung, Sutton, Gehrmann, Schuh, Shi, Tsvyashchenko, Maynez,
  Rao, Barnes, Tay, Shazeer, Prabhakaran, Reif, Du, Hutchinson, Pope, Bradbury,
  Austin, Isard, Gur{-}Ari, Yin, Duke, Levskaya, Ghemawat, Dev, Michalewski,
  Garcia, Misra, Robinson, Fedus, Zhou, Ippolito, Luan, Lim, Zoph, Spiridonov,
  Sepassi, Dohan, Agrawal, Omernick, Dai, Pillai, Pellat, Lewkowycz, Moreira,
  Child, Polozov, Lee, Zhou, Wang, Saeta, Diaz, Firat, Catasta, Wei,
  Meier{-}Hellstern, Eck, Dean, Petrov, and Fiedel}]{palm}
Aakanksha Chowdhery, Sharan Narang, Jacob Devlin, Maarten Bosma, Gaurav Mishra,
  Adam Roberts, Paul Barham, Hyung~Won Chung, Charles Sutton, Sebastian
  Gehrmann, Parker Schuh, Kensen Shi, Sasha Tsvyashchenko, Joshua Maynez,
  Abhishek Rao, Parker Barnes, Yi~Tay, Noam Shazeer, Vinodkumar Prabhakaran,
  Emily Reif, Nan Du, Ben Hutchinson, Reiner Pope, James Bradbury, Jacob
  Austin, Michael Isard, Guy Gur{-}Ari, Pengcheng Yin, Toju Duke, Anselm
  Levskaya, Sanjay Ghemawat, Sunipa Dev, Henryk Michalewski, Xavier Garcia,
  Vedant Misra, Kevin Robinson, Liam Fedus, Denny Zhou, Daphne Ippolito, David
  Luan, Hyeontaek Lim, Barret Zoph, Alexander Spiridonov, Ryan Sepassi, David
  Dohan, Shivani Agrawal, Mark Omernick, Andrew~M. Dai,
  Thanumalayan~Sankaranarayana Pillai, Marie Pellat, Aitor Lewkowycz, Erica
  Moreira, Rewon Child, Oleksandr Polozov, Katherine Lee, Zongwei Zhou, Xuezhi
  Wang, Brennan Saeta, Mark Diaz, Orhan Firat, Michele Catasta, Jason Wei,
  Kathy Meier{-}Hellstern, Douglas Eck, Jeff Dean, Slav Petrov, and Noah
  Fiedel. 2022.
\newblock Palm: Scaling language modeling with pathways.
\newblock \emph{CoRR}, abs/2204.02311.

\bibitem[{Chung et~al.(2022)Chung, Hou, Longpre, Zoph, Tay, Fedus, Li, Wang,
  Dehghani, Brahma, Webson, Gu, Dai, Suzgun, Chen, Chowdhery, Castro-Ros,
  Pellat, Robinson, Valter, Narang, Mishra, Yu, Zhao, Huang, Dai, Yu, Petrov,
  Chi, Dean, Devlin, Roberts, Zhou, Le, and Wei}]{flant5}
Hyung~Won Chung, Le~Hou, Shayne Longpre, Barret Zoph, Yi~Tay, William Fedus,
  Yunxuan Li, Xuezhi Wang, Mostafa Dehghani, Siddhartha Brahma, Albert Webson,
  Shixiang~Shane Gu, Zhuyun Dai, Mirac Suzgun, Xinyun Chen, Aakanksha
  Chowdhery, Alex Castro-Ros, Marie Pellat, Kevin Robinson, Dasha Valter,
  Sharan Narang, Gaurav Mishra, Adams Yu, Vincent Zhao, Yanping Huang, Andrew
  Dai, Hongkun Yu, Slav Petrov, Ed~H. Chi, Jeff Dean, Jacob Devlin, Adam
  Roberts, Denny Zhou, Quoc~V. Le, and Jason Wei. 2022.
\newblock \href {https://doi.org/10.48550/ARXIV.2210.11416} {Scaling
  instruction-finetuned language models}.

\bibitem[{Clark et~al.(2020)Clark, Luong, Le, and Manning}]{clark2020electra}
Kevin Clark, Minh-Thang Luong, Quoc~V Le, and Christopher~D Manning. 2020.
\newblock Electra: Pre-training text encoders as discriminators rather than
  generators.
\newblock \emph{arXiv preprint arXiv:2003.10555}.

\bibitem[{De~Marneffe et~al.(2019)De~Marneffe, Simons, and
  Tonhauser}]{de2019_CB}
Marie-Catherine De~Marneffe, Mandy Simons, and Judith Tonhauser. 2019.
\newblock The commitmentbank: Investigating projection in naturally occurring
  discourse.
\newblock In \emph{proceedings of Sinn und Bedeutung}, volume~23, pages
  107--124.

\bibitem[{Devlin et~al.(2019)Devlin, Chang, Lee, and
  Toutanova}]{devlin2018bert}
Jacob Devlin, Ming-Wei Chang, Kenton Lee, and Kristina Toutanova. 2019.
\newblock {BERT}: Pre-training of deep bidirectional transformers for language
  understanding.
\newblock In \emph{Proceedings of the 2019 Conference of the North {A}merican
  Chapter of the Association for Computational Linguistics: Human Language
  Technologies, Volume 1 (Long and Short Papers)}, pages 4171--4186,
  Minneapolis, Minnesota. Association for Computational Linguistics.

\bibitem[{Du et~al.(2022)Du, Huang, Dai, Tong, Lepikhin, Xu, Krikun, Zhou, Yu,
  Firat, Zoph, Fedus, Bosma, Zhou, Wang, Wang, Webster, Pellat, Robinson,
  Meier{-}Hellstern, Duke, Dixon, Zhang, Le, Wu, Chen, and Cui}]{glam}
Nan Du, Yanping Huang, Andrew~M. Dai, Simon Tong, Dmitry Lepikhin, Yuanzhong
  Xu, Maxim Krikun, Yanqi Zhou, Adams~Wei Yu, Orhan Firat, Barret Zoph, Liam
  Fedus, Maarten~P. Bosma, Zongwei Zhou, Tao Wang, Yu~Emma Wang, Kellie
  Webster, Marie Pellat, Kevin Robinson, Kathleen~S. Meier{-}Hellstern, Toju
  Duke, Lucas Dixon, Kun Zhang, Quoc~V. Le, Yonghui Wu, Zhifeng Chen, and
  Claire Cui. 2022.
\newblock Glam: Efficient scaling of language models with mixture-of-experts.
\newblock In \emph{{ICML}}, volume 162 of \emph{Proceedings of Machine Learning
  Research}, pages 5547--5569. {PMLR}.

\bibitem[{Goodfellow et~al.(2014)Goodfellow, Pouget-Abadie, Mirza, Xu,
  Warde-Farley, Ozair, Courville, and Bengio}]{goodfellow2014generative}
Ian Goodfellow, Jean Pouget-Abadie, Mehdi Mirza, Bing Xu, David Warde-Farley,
  Sherjil Ozair, Aaron Courville, and Yoshua Bengio. 2014.
\newblock Generative adversarial nets.
\newblock \emph{Advances in neural information processing systems}, 27.

\bibitem[{He et~al.(2021{\natexlab{a}})He, Zhou, Ma, Berg-Kirkpatrick, and
  Neubig}]{he2021towards}
Junxian He, Chunting Zhou, Xuezhe Ma, Taylor Berg-Kirkpatrick, and Graham
  Neubig. 2021{\natexlab{a}}.
\newblock Towards a unified view of parameter-efficient transfer learning.
\newblock \emph{arXiv preprint arXiv:2110.04366}.

\bibitem[{He et~al.(2021{\natexlab{b}})He, Gao, and Chen}]{debertav3}
Pengcheng He, Jianfeng Gao, and Weizhu Chen. 2021{\natexlab{b}}.
\newblock Debertav3: Improving deberta using electra-style pre-training with
  gradient-disentangled embedding sharing.
\newblock \emph{CoRR}, abs/2111.09543.

\bibitem[{Levesque et~al.(2012)Levesque, Davis, and Morgenstern}]{WSC2012}
Hector Levesque, Ernest Davis, and Leora Morgenstern. 2012.
\newblock The winograd schema challenge.
\newblock In \emph{Thirteenth International Conference on the Principles of
  Knowledge Representation and Reasoning}. Citeseer.

\bibitem[{Liu et~al.(2021{\natexlab{a}})Liu, Zheng, Du, Ding, Qian, Yang, and
  Tang}]{ptuning-paper}
Xiao Liu, Yanan Zheng, Zhengxiao Du, Ming Ding, Yujie Qian, Zhilin Yang, and
  Jie Tang. 2021{\natexlab{a}}.
\newblock {GPT} understands, too.
\newblock \emph{CoRR}, abs/2103.10385.

\bibitem[{Liu et~al.(2021{\natexlab{b}})Liu, Zheng, Du, Ding, Qian, Yang, and
  Tang}]{liu2021gpt}
Xiao Liu, Yanan Zheng, Zhengxiao Du, Ming Ding, Yujie Qian, Zhilin Yang, and
  Jie Tang. 2021{\natexlab{b}}.
\newblock Gpt understands, too.
\newblock \emph{arXiv preprint arXiv:2103.10385}.

\bibitem[{Liu et~al.(2019)Liu, Ott, Goyal, Du, Joshi, Chen, Levy, Lewis,
  Zettlemoyer, and Stoyanov}]{liu2019roberta}
Yinhan Liu, Myle Ott, Naman Goyal, Jingfei Du, Mandar Joshi, Danqi Chen, Omer
  Levy, Mike Lewis, Luke Zettlemoyer, and Veselin Stoyanov. 2019.
\newblock \href {http://arxiv.org/abs/1907.11692} {Ro{BERT}a: A robustly
  optimized bert pretraining approach}.

\bibitem[{Logan~IV et~al.(2021)Logan~IV, Bala{\v{z}}evi{\'c}, Wallace, Petroni,
  Singh, and Riedel}]{logan2021cutting}
Robert~L Logan~IV, Ivana Bala{\v{z}}evi{\'c}, Eric Wallace, Fabio Petroni,
  Sameer Singh, and Sebastian Riedel. 2021.
\newblock Cutting down on prompts and parameters: Simple few-shot learning with
  language models.
\newblock \emph{arXiv preprint arXiv:2106.13353}.

\bibitem[{Mahabadi et~al.(2022)Mahabadi, Zettlemoyer, Henderson, Mathias,
  Saeidi, Stoyanov, and Yazdani}]{mahabadi2022prompt}
Rabeeh~Karimi Mahabadi, Luke Zettlemoyer, James Henderson, Lambert Mathias,
  Marzieh Saeidi, Veselin Stoyanov, and Majid Yazdani. 2022.
\newblock Prompt-free and efficient few-shot learning with language models.
\newblock In \emph{Proceedings of the 60th Annual Meeting of the Association
  for Computational Linguistics (Volume 1: Long Papers)}, pages 3638--3652.

\bibitem[{Meng et~al.(2022)Meng, Huang, Zhang, and Han}]{meng2022generating}
Yu~Meng, Jiaxin Huang, Yu~Zhang, and Jiawei Han. 2022.
\newblock Generating training data with language models: Towards zero-shot
  language understanding.
\newblock \emph{arXiv preprint arXiv:2202.04538}.

\bibitem[{Mostafazadeh et~al.(2017)Mostafazadeh, Roth, Louis, Chambers, and
  Allen}]{story_cloze}
Nasrin Mostafazadeh, Michael Roth, Annie Louis, Nathanael Chambers, and James
  Allen. 2017.
\newblock Lsdsem 2017 shared task: The story cloze test.
\newblock In \emph{Proceedings of the 2nd Workshop on Linking Models of
  Lexical, Sentential and Discourse-level Semantics}, pages 46--51.

\bibitem[{Nie et~al.(2020)Nie, Williams, Dinan, Bansal, Weston, and
  Kiela}]{NieWDBWK20_ANLI}
Yixin Nie, Adina Williams, Emily Dinan, Mohit Bansal, Jason Weston, and Douwe
  Kiela. 2020.
\newblock Adversarial {NLI:} {A} new benchmark for natural language
  understanding.
\newblock In \emph{{ACL}}, pages 4885--4901. Association for Computational
  Linguistics.

\bibitem[{Oord et~al.(2018)Oord, Li, and Vinyals}]{oord2018representation}
Aaron van~den Oord, Yazhe Li, and Oriol Vinyals. 2018.
\newblock Representation learning with contrastive predictive coding.
\newblock \emph{arXiv preprint arXiv:1807.03748}.

\bibitem[{Pilehvar and Camacho{-}Collados(2018)}]{wic-paper}
Mohammad~Taher Pilehvar and Jos{\'{e}} Camacho{-}Collados. 2018.
\newblock Wic: 10, 000 example pairs for evaluating context-sensitive
  representations.
\newblock \emph{CoRR}, abs/1808.09121.

\bibitem[{Radford et~al.(2018)Radford, Wu, Child, Luan, Amodei, and
  Sutskever}]{radford2018gpt}
Alec Radford, Jeffrey Wu, Rewon Child, David Luan, Dario Amodei, and Ilya
  Sutskever. 2018.
\newblock \href
  {https://d4mucfpksywv.cloudfront.net/better-language-models/language-models.pdf}
  {Language models are unsupervised multitask learners}.

\bibitem[{Rae et~al.(2021)Rae, Borgeaud, Cai, Millican, Hoffmann, Song,
  Aslanides, Henderson, Ring, Young et~al.}]{rae2021scaling}
Jack~W Rae, Sebastian Borgeaud, Trevor Cai, Katie Millican, Jordan Hoffmann,
  Francis Song, John Aslanides, Sarah Henderson, Roman Ring, Susannah Young,
  et~al. 2021.
\newblock Scaling language models: Methods, analysis \& insights from training
  gopher.
\newblock \emph{arXiv preprint arXiv:2112.11446}.

\bibitem[{Raffel et~al.(2019)Raffel, Shazeer, Roberts, Lee, Narang, Matena,
  Zhou, Li, and Liu}]{T5-paper}
Colin Raffel, Noam Shazeer, Adam Roberts, Katherine Lee, Sharan Narang, Michael
  Matena, Yanqi Zhou, Wei Li, and Peter~J. Liu. 2019.
\newblock Exploring the limits of transfer learning with a unified text-to-text
  transformer.
\newblock \emph{CoRR}, abs/1910.10683.

\bibitem[{Roemmele et~al.(2011)Roemmele, Bejan, and Gordon}]{COPA2011}
Melissa Roemmele, Cosmin~Adrian Bejan, and Andrew~S Gordon. 2011.
\newblock Choice of plausible alternatives: An evaluation of commonsense causal
  reasoning.
\newblock In \emph{AAAI Spring Symposium: Logical Formalizations of Commonsense
  Reasoning}, pages 90--95.

\bibitem[{Sakaguchi et~al.(2020)Sakaguchi, Bras, Bhagavatula, and
  Choi}]{SakaguchiBBC20_winogrande}
Keisuke Sakaguchi, Ronan~Le Bras, Chandra Bhagavatula, and Yejin Choi. 2020.
\newblock Winogrande: An adversarial winograd schema challenge at scale.
\newblock In \emph{{AAAI}}, pages 8732--8740. {AAAI} Press.

\bibitem[{Sanh et~al.(2021)Sanh, Webson, Raffel, Bach, Sutawika, Alyafeai,
  Chaffin, Stiegler, Scao, Raja, Dey, Bari, Xu, Thakker, Sharma, Szczechla,
  Kim, Chhablani, Nayak, Datta, Chang, Jiang, Wang, Manica, Shen, Yong, Pandey,
  Bawden, Wang, Neeraj, Rozen, Sharma, Santilli, F{\'{e}}vry, Fries, Teehan,
  Biderman, Gao, Bers, Wolf, and Rush}]{T0-paper}
Victor Sanh, Albert Webson, Colin Raffel, Stephen~H. Bach, Lintang Sutawika,
  Zaid Alyafeai, Antoine Chaffin, Arnaud Stiegler, Teven~Le Scao, Arun Raja,
  Manan Dey, M.~Saiful Bari, Canwen Xu, Urmish Thakker, Shanya Sharma, Eliza
  Szczechla, Taewoon Kim, Gunjan Chhablani, Nihal~V. Nayak, Debajyoti Datta,
  Jonathan Chang, Mike~Tian{-}Jian Jiang, Han Wang, Matteo Manica, Sheng Shen,
  Zheng~Xin Yong, Harshit Pandey, Rachel Bawden, Thomas Wang, Trishala Neeraj,
  Jos Rozen, Abheesht Sharma, Andrea Santilli, Thibault F{\'{e}}vry, Jason~Alan
  Fries, Ryan Teehan, Stella Biderman, Leo Gao, Tali Bers, Thomas Wolf, and
  Alexander~M. Rush. 2021.
\newblock Multitask prompted training enables zero-shot task generalization.
\newblock \emph{CoRR}, abs/2110.08207.

\bibitem[{Schick and Sch{\"{u}}tze(2020)}]{PET-paper}
Timo Schick and Hinrich Sch{\"{u}}tze. 2020.
\newblock It's not just size that matters: Small language models are also
  few-shot learners.
\newblock \emph{CoRR}, abs/2009.07118.

\bibitem[{Srivastava et~al.(2022)Srivastava, Rastogi, Rao, Shoeb, Abid, Fisch,
  Brown, Santoro, Gupta, Garriga-Alonso et~al.}]{bigbench}
Aarohi Srivastava, Abhinav Rastogi, Abhishek Rao, Abu Awal~Md Shoeb, Abubakar
  Abid, Adam Fisch, Adam~R Brown, Adam Santoro, Aditya Gupta, Adri{\`a}
  Garriga-Alonso, et~al. 2022.
\newblock Beyond the imitation game: Quantifying and extrapolating the
  capabilities of language models.
\newblock \emph{arXiv preprint arXiv:2206.04615}.

\bibitem[{Suzgun et~al.(2022)Suzgun, Scales, Schärli, Gehrmann, Tay, Chung,
  Chowdhery, Le, Chi, Zhou, and Wei}]{bbh}
Mirac Suzgun, Nathan Scales, Nathanael Schärli, Sebastian Gehrmann, Yi~Tay,
  Hyung~Won Chung, Aakanksha Chowdhery, Quoc~V. Le, Ed~H. Chi, Denny Zhou, and
  Jason Wei. 2022.
\newblock \href {https://doi.org/10.48550/ARXIV.2210.09261} {Challenging
  big-bench tasks and whether chain-of-thought can solve them}.

\bibitem[{Tay et~al.(2022)Tay, Dehghani, Tran, Garcia, Bahri, Schuster, Zheng,
  Houlsby, and Metzler}]{ul2}
Yi~Tay, Mostafa Dehghani, Vinh~Q. Tran, Xavier Garcia, Dara Bahri, Tal
  Schuster, Huaixiu~Steven Zheng, Neil Houlsby, and Donald Metzler. 2022.
\newblock \href {https://doi.org/10.48550/ARXIV.2205.05131} {Unifying language
  learning paradigms}.

\bibitem[{Wang et~al.(2019{\natexlab{a}})Wang, Pruksachatkun, Nangia, Singh,
  Michael, Hill, Levy, and Bowman}]{wang2019superglue}
Alex Wang, Yada Pruksachatkun, Nikita Nangia, Amanpreet Singh, Julian Michael,
  Felix Hill, Omer Levy, and Samuel~R Bowman. 2019{\natexlab{a}}.
\newblock Superglue: A stickier benchmark for general-purpose language
  understanding systems.
\newblock \emph{arXiv preprint arXiv:1905.00537}.

\bibitem[{Wang et~al.(2019{\natexlab{b}})Wang, Singh, Michael, Hill, Levy, and
  Bowman}]{wang2019glue}
Alex Wang, Amanpreet Singh, Julian Michael, Felix Hill, Omer Levy, and
  Samuel~R. Bowman. 2019{\natexlab{b}}.
\newblock {GLUE}: A multi-task benchmark and analysis platform for natural
  language understanding.
\newblock In the Proceedings of ICLR.

\bibitem[{Wang et~al.(2022)Wang, Mishra, Alipoormolabashi, Kordi, Mirzaei,
  Arunkumar, Ashok, Dhanasekaran, Naik, Stap, Pathak, Karamanolakis, Lai,
  Purohit, Mondal, Anderson, Kuznia, Doshi, Patel, Pal, Moradshahi, Parmar,
  Purohit, Varshney, Kaza, Verma, Puri, Karia, Sampat, Doshi, Mishra, Reddy,
  Patro, Dixit, Shen, Baral, Choi, Smith, Hajishirzi, and Khashabi}]{1600tasks}
Yizhong Wang, Swaroop Mishra, Pegah Alipoormolabashi, Yeganeh Kordi, Amirreza
  Mirzaei, Anjana Arunkumar, Arjun Ashok, Arut~Selvan Dhanasekaran, Atharva
  Naik, David Stap, Eshaan Pathak, Giannis Karamanolakis, Haizhi~Gary Lai,
  Ishan Purohit, Ishani Mondal, Jacob Anderson, Kirby Kuznia, Krima Doshi,
  Maitreya Patel, Kuntal~Kumar Pal, Mehrad Moradshahi, Mihir Parmar, Mirali
  Purohit, Neeraj Varshney, Phani~Rohitha Kaza, Pulkit Verma, Ravsehaj~Singh
  Puri, Rushang Karia, Shailaja~Keyur Sampat, Savan Doshi, Siddhartha Mishra,
  Sujan Reddy, Sumanta Patro, Tanay Dixit, Xudong Shen, Chitta Baral, Yejin
  Choi, Noah~A. Smith, Hannaneh Hajishirzi, and Daniel Khashabi. 2022.
\newblock \href {https://doi.org/10.48550/ARXIV.2204.07705} {Benchmarking
  generalization via in-context instructions on 1,600+ language tasks}.

\bibitem[{Wei et~al.(2021)Wei, Bosma, Zhao, Guu, Yu, Lester, Du, Dai, and
  Le}]{FLAN}
Jason Wei, Maarten Bosma, Vincent~Y. Zhao, Kelvin Guu, Adams~Wei Yu, Brian
  Lester, Nan Du, Andrew~M. Dai, and Quoc~V. Le. 2021.
\newblock Finetuned language models are zero-shot learners.
\newblock \emph{CoRR}, abs/2109.01652.

\bibitem[{Xia et~al.(2022)Xia, Artetxe, Du, Chen, and
  Stoyanov}]{xia2022prompting}
Mengzhou Xia, Mikel Artetxe, Jingfei Du, Danqi Chen, and Ves Stoyanov. 2022.
\newblock Prompting electra: Few-shot learning with discriminative pre-trained
  models.
\newblock \emph{arXiv preprint arXiv:2205.15223}.

\bibitem[{Xiong et~al.(2019)Xiong, Du, Wang, and
  Stoyanov}]{xiong2019pretrained}
Wenhan Xiong, Jingfei Du, William~Yang Wang, and Veselin Stoyanov. 2019.
\newblock Pretrained encyclopedia: Weakly supervised knowledge-pretrained
  language model.
\newblock \emph{arXiv preprint arXiv:1912.09637}.

\bibitem[{Yao et~al.(2022)Yao, Dong, Zhang, Zhang, Xie, Liu, Lin, Sun, and
  Wang}]{yao2022prompt}
Yuan Yao, Bowen Dong, Ao~Zhang, Zhengyan Zhang, Ruobing Xie, Zhiyuan Liu, Leyu
  Lin, Maosong Sun, and Jianyong Wang. 2022.
\newblock Prompt tuning for discriminative pre-trained language models.
\newblock \emph{arXiv preprint arXiv:2205.11166}.

\bibitem[{Zellers et~al.(2019)Zellers, Holtzman, Bisk, Farhadi, and
  Choi}]{ZellersHBFC19_hellaswag}
Rowan Zellers, Ari Holtzman, Yonatan Bisk, Ali Farhadi, and Yejin Choi. 2019.
\newblock Hellaswag: Can a machine really finish your sentence?
\newblock In \emph{{ACL} {(1)}}, pages 4791--4800. Association for
  Computational Linguistics.

\end{thebibliography}

\newpage
\onecolumn
\appendix
\section{Examples of Minimal Prompt}
Here we provide Table~\ref{tab:task_formulate_example} for some examples of how to construct minimal prompted data according to \S~\ref{sec:unifydiscriminativetasks}.

\begin{table*}[htbp]
  \centering
    \resizebox{1.0\textwidth}{!}{\begin{tabular}{c|c|p{37em}|c}
    \toprule
    \multirow{2}[2]{*}{\textbf{Category}} & \multirow{2}[2]{*}{\textbf{Task Type}} & \multicolumn{1}{c|}{\multirow{2}[2]{*}{\textbf{Our Minmal Prompt}}} & \multirow{2}[2]{*}{\textbf{Label}} \\
          &       & \multicolumn{1}{c|}{} &  \\
    \midrule
    \multirow{4}[8]{*}{yes/no} & \multirow{2}[4]{*}{\shortstack{Paraphrase \\ Identification}} & John is Lily's husband. Lily is John's wife & 1 \\
\cmidrule{3-4}          &       & John is Lily's husband. Lily is John's mother. & 0 \\
\cmidrule{2-4}          & \multicolumn{1}{c|}{\multirow{4}[4]{*}{\shortstack{Natural \\ Language  \\ Inference}}} & \underline{Premise:} Dana Reeve, the widow of the actor Christopher Reeve, has died of lung cancer at age 44. \underline{Hypothesis:} Dana Reeve had an accident. & 1 \\
\cmidrule{3-4}          &       & \underline{Premise:} Dana Reeve, the widow of the actor Christopher Reeve, has died of lung cancer at age 44. \underline{Hypothesis:} Christopher Reeve had an accident. & 0 \\
    \midrule
    \multirow{10}[20]{*}{multi-choice} & \multirow{2}[4]{*}{\shortstack{Coreference \\ Resolution}} & Jane gives Joan candy because Joan was hungry. & 1 \\
\cmidrule{3-4}          &       & Jane gives Joan candy because Jane was hungry. & 0 \\
\cmidrule{2-4}          
& \multirow{2}[4]{*}{\shortstack{Question \\ Answer}} & The earth moves around the sun. What is the earch to the sun? Planet & 1 \\
\cmidrule{3-4}          &       & The earth moves around the sun. What is the earch to the sun? Satellite & 0 \\
\cmidrule{2-4}          & \multirow{2}[4]{*}{\shortstack{Topic \\ Classification}} & Open Source Apps Developer SugarCRM Releases Sugar.Sales 1.1. Science and technology & 1 \\
\cmidrule{3-4}          &       & Open Source Apps Developer SugarCRM Releases Sugar.Sales 1.1. Sports & 0 \\
\cmidrule{2-4}          & \multirow{2}[4]{*}{\shortstack{Sentence \\ Completion}} & A boy is running down a track. The boy lifts his body above the height of a pole. & 1 \\
\cmidrule{3-4}          &       & A boy is running down a track. The boy stands on his hands and springs. & 0 \\
\cmidrule{2-4}          & \multirow{2}[4]{*}{\shortstack{Sentiment \\ Classification}} & I really love this movie. Positive & 1 \\
\cmidrule{3-4}          &       & I don't like this movie. Negative & 1 \\
\bottomrule
    \end{tabular}}%
\caption{Examples of how we unify discriminative tasks. The underlined text represents additional words not present in raw inputs. Note that this is just our implementation of the UD formulation and there can be other ways of task formulation under the UD framework. Some tasks can either be yes/no tasks or multi-choice tasks, depending on how options are provided.}
\label{tab:task_formulate_example}
\end{table*}%

\section{Full Experiment Results}

\subsection{Evaluation on Big-Bench}\label{sec:bigbench}
Here we report the full results for 13 tasks in the Big-Bench \citet{bigbench}, which is also utilized in original T0 paper~\cite{T0-paper}. All the tasks from BIG-Bench are ensured unseen in our training set for the zero-shot setting. The results are displayed in Table ~\ref{tab:bigbench}, where UD outperforms T0 by 4-8 points on different scales.

\begin{table*}
\resizebox{\textwidth}{!}{%
    \begin{tabular}{l|ccccccccccccc|c}
    \toprule[1pt]
    \multirow{2}{*}{Model} 
        & \multirow{2}{*}{\shortstack{code \\ desc.}}
        & \multirow{2}{*}{\shortstack{conce\\-ptual}}
        & \multirow{2}{*}{\shortstack{known\\unknowns}}
        & \multirow{2}{*}{\shortstack{logic \\ grid}}
        & \multirow{2}{*}{\shortstack{logic \\ deduction}}
        & \multirow{2}{*}{\shortstack{miscon\\-ceptions}}
        & \multirow{2}{*}{\shortstack{novel\\concepts}}
        & \multirow{2}{*}{\shortstack{strate\\-gyqa}}
        & \multirow{2}{*}{\shortstack{wino\\-why}}
        & \multirow{2}{*}{\shortstack{syllo\\-gisms}}
        & \multirow{2}{*}{\shortstack{movie\\dialog}}
        & \multirow{2}{*}{\shortstack{lang\\-uage\_id}}
        & \multirow{2}{*}{\shortstack{vita\\-minc}} 
        & \multirow{2}{*}{Avg.}\\
    &&&&&&&&&&&&&&\\
    \midrule[1pt]
    UD-DeBERTaV3 & 76.7 & 64.1 & 76.1 & 39.9 & 54.9 & 50.2 & 50.0 & 59.9 & 45.8 & 50.4 & 57.7 & 13.3 & 61.5 & 53.9 \\
    \midrule[1pt]
    T0-Large$\star$ & 14.1 & 40.4 & \textbf{60.4} & \textbf{38.0} & \textbf{41.2} & 50.0 & 10.0 & \textbf{52.3} & \textbf{49.7} & 50.3 & 46.8 & 16.0 & 46.2 & 39.6 \\
    UD-Large & \textbf{51.7} & \textbf{54.4} & 47.8 & 33.4 & 34.6 & \textbf{50.2} & \textbf{26.5} & 47.0 & 45.7 & \textbf{50.6} & \textbf{51.7} & \textbf{16.3} & \textbf{55.8} & \textbf{43.5} \\
    
    \midrule[1pt]
    T0-XL$\star$ & 23.4 & 48.1 & 64.6 & \textbf{42.5} & \textbf{50.1} & \textbf{52.7} & 25.0 & 53.1 & \textbf{45.4} & 50.2 & 47.7 & 19.0 & 60.0 & 44.8 \\
    UD-XL & \textbf{53.3} & \textbf{73.8} & \textbf{65.2} & 37.2 & 37.8 & 48.0 & \textbf{35.3} & \textbf{53.1} & 45.3 & \textbf{50.4} & \textbf{50.1} & \textbf{22.9} & \textbf{63.7} & \textbf{48.9} \\
    \midrule[1pt]
    T0-XXL$\dagger$ & 36.7 & 62.5 & 63.0 & 39.6 & 55.4 & \textbf{52.5} & 15.6 & 52.7 & 47.4 & \textbf{51.8} & 53.8 & 20.7 & 64.7 & 47.4\\
    UD-XXL & 61.7 & 71.8 & 76.1 & 38.0 & 59.1 & 49.3 & \textbf{61.8} & 61.3 & 45.9 & 50.1 & 57.3 & 21.6 & 67.2 & 55.5 \\
    UD+-XXL & \textbf{63.3} & \textbf{82.5} & \textbf{84.8} & \textbf{39.2} & \textbf{67.5} & 49.3 & 58.8 & \textbf{64.2} & \textbf{47.5} & 50.4 & \textbf{57.9} & \textbf{27.3} & \textbf{70.2} & \textbf{58.7} \\
    \bottomrule[1pt]
    
    \end{tabular}%
    }
\caption{Zero-shot performance of Universal Discriminator and T0 on Big-Bench test tasks used in T0 paper. Results with $\dagger$ are reported by~\citeauthor{T0-paper}, and results with $\star$ are reproduced in our framework.}
\label{tab:bigbench}
\end{table*}

\subsection{Evaluation on BBH}\label{sec:bbh}
Here we report the full results for 22 discriminative tasks from BBH \citep{bbh}. For reference, we reproduce Flan-T5\citep{flant5}'s zero-shot performance on BBH tasks by evaluating their public checkpoints. All the tasks from BBH are ensured unseen in our training set for the zero-shot setting. The results are displayed in Table~\ref{tab:bbh_full}, where UD constantly performs better than T0 and Flan-T5 on all the scales even though Flan-T5 is trained on a much broader scope of tasks than UD is.

\begin{table*}
\resizebox{\textwidth}{!}{%

    \begin{tabular}{l|ccc|ccc|cccc}
    \toprule[1pt]
    Dataset & T0-Large & Flan-T5-Large & UD-Large & T0-XL & Flan-T5-XL & UD-XL & T0-XXL & Flan-T5-XXL & UD-XXL & UD+-XXL \\
    \midrule[1pt]
    boolean\_expression  & 48.4 & 49.6 & \textbf{64.0} & 47.6 & 54.8 & \textbf{68.4} & 46.4 & 56.8 & \textbf{68.4} & 66.0 \\
    causal\_judgement & 56.2 & 59.4 & \textbf{61.5} & 58.8 & 59.9 & \textbf{63.6} & 62.0 & 60.9 & \textbf{65.2} & 63.6 \\
    data\_understanding & 30.4 & 18.8 & \textbf{30.4} & 38.8 & 34.8 & \textbf{41.2} & \textbf{63.2} & 56.8 & 51.6 & 53.2  \\
    disambiguation\_qa & 54.4 & 34.8 & \textbf{68.4} & 61.2 & \textbf{66.8} & 65.2 & 64.4 & 66.8 & \textbf{67.2} & 66.8 \\
    formal\_fallacies & 54.4 & \textbf{55.6} & 50.4 & 52.4 & \textbf{54.0} & 46.4 & 52.0 & 55.2 & 54.0 & \textbf{58.8} \\
    geometric\_shapes & 0.0 & \textbf{21.6} & 9.6 & 0.0 & \textbf{20.0} & 9.6 & 11.2 & \textbf{31.2} & 9.6 & 9.6 \\
    hyperbaton & \textbf{72.0} & 59.6 & 71.2 & 52.4 & 58.8 & \textbf{66.8} & 63.2 & 70.8 & 68.0 & \textbf{82.0}\\
    logical\_deduction\_five\_objects & 34.8 & \textbf{40.0} & 32.8 & 38.8 & \textbf{48.0} & 39.2 & 46.4 & 53.6 & 58.4 & \textbf{65.2} \\
    logical\_deduction\_seven\_objects & 27.6 & \textbf{40.4} & 25.2 & 37.6 & \textbf{52.4} & 32.0 & 50.4 & 60.0 & 56.4 & \textbf{67.2}  \\
    logical\_deduction\_three\_objects & 49.2 & 37.6 & \textbf{60.4} & 62.8 & 64.8 & \textbf{69.2} & 65.6 & 74.4 & 80.8 & \textbf{83.2} \\
    movie\_recommendation & 51.4 & 55.0 & \textbf{60.4} & 55.0 & 47.4 & \textbf{69.6} & 61.0 & 38.5 & 73.2 & \textbf{78.8} \\
    navigate & 58.8 & 56.4 & \textbf{63.6} & 60.4 & 59.2 & 58.4 & 65.6 & 60.8 & 63.2 & 64.8 \\
    penguins\_in\_a\_table & 36.3 & 32.9 & \textbf{36.3} & 34.3 & \textbf{42.5} & 41.1 & 40.4 & 41.1 & 39.7 & \textbf{46.6} \\
    reasoning\_about\_colored\_objects & 39.2 & \textbf{40.4} & 36.4 & 41.6 & 47.2 & \textbf{54.4} & 56.8 & 61.6 & 57.2 & \textbf{63.2}\\
    ruin\_names & 23.0 & 22.6 & \textbf{44.4} & 21.8 & \textbf{33.5} & 24.4 & 17.8 & 34.7 & 35.6 & \textbf{68.8} \\
    snarks & 48.3 & 56.1 & \textbf{74.7} & 45.5 & 55.6 & \textbf{73.0} & 55.1 & 72.5 & 75.3 & \textbf{82.0} \\
    sports\_understanding & 53.2 & \textbf{55.6} & 54.8 & 47.6 & \textbf{52.4} & 51.6 & 52.8 & \textbf{60.0} & 57.6 & 56.0 \\
    temporal\_sequences & 13.2 & \textbf{25.2} & 23.6 & 24.8 & 22.4 & \textbf{63.2} & 14.8 & 28.8 & 43.2 & \textbf{60.8} \\
    tracking\_shuffled\_objects\_five\_objects & \textbf{12.8} & 12.4 & 12.0 & 12.8 & 12.0 & \textbf{13.2} & 12.0 & 15.2 & 12.4 & \textbf{20.0}  \\
    tracking\_shuffled\_objects\_seven\_objects & 7.6 & 8.4 & \textbf{9.6} & 8.8 & \textbf{9.2} & 8.4 & 8.0 & 13.2 & 8.4 & \textbf{14.0} \\
    tracking\_shuffled\_objects\_three\_objects & 33.2 & \textbf{33.6} & 31.2 & 33.6 & 32.8 & \textbf{34.8} & 29.6 & 24.4 & \textbf{33.6} & 20.8  \\
    web\_of\_lies & 51.2 & \textbf{52.4} & 51.2 & 51.2 & \textbf{52.4} & 47.6 & 50.8 & 50.0 & 50.4 & \textbf{56.8} \\
    \midrule[1pt]
    Avg.  & 38.9 & 39.5 & \textbf{44.2} & 40.4 & 44.6 & \textbf{47.3} & 45.0 & 49.4 & 51.3 & \textbf{56.7} \\
    \bottomrule[1pt]
    \end{tabular}%
}
  \caption{Zero-shot performance of Universal Discriminator, T0, and Flan-T5 on BBH test tasks \citep{bbh}.}
  \label{tab:bbh_full}%
\end{table*}%

\section{More Ablation Studies}

\subsection{Ablation on Base Models}\label{sec:base_models}
\begin{table*}[ht]
\setlength{\tabcolsep}{0.9mm}
\centering
\resizebox{\textwidth}{!}{%
    \begin{tabular}{l|l|ccccc|ccc|cc|c|c}
        \toprule[1pt]
        & \multirow{2}*{Base Model}
        & \multicolumn{5}{c|}{\textbf{Natural Language Inference}} & \multicolumn{3}{|c|}{\textbf{Sentence Completion}} & \multicolumn{2}{c|}{\textbf{Coreference}} & \multicolumn{1}{c|}{\textbf{WSD}} & \multirow{2}{*}{Avg.} \\
    & & RTE & CB & ANLI1 & ANLI2 & ANLI3 & COPA & Hella. & Story. & WSC & Wino. & WiC &  \\
    \midrule[1pt]
    \multirow{2}*{\shortstack{Encoder}}
    & DeBERTa-V3 (304M) 
        & 71.1
        & 76.8
        & 43.8
        & 41.3
        & 45.7
        & 96.0
        & 60.7
        & 97.4
        & 66.4
        & 83.6
        & 53.3
        & 66.9 \\
    & DeBERTa-V2 (1.5B) 
        & 77.6
        & 80.4
        & 43.2
        & 39.3
        & 44.8
        & 95.0
        & 67.2
        & 98.2
        & 74.0	& 82.1 & 56.0	& 68.9\\ \midrule
    \multirow{2}*{\shortstack{Enc-Dec}} & T5-Encoder (400M) 
        & 75.1	& 55.5	& 32.9	& 32.3	& 33.7	& 84.6	& 28.2	& 94.0	& 63.0	& 54.6	& 51.2	& 55.0 \\
    & T5-Encoder (1.5B)  & 79.7	& 68.9	& 43.1	& 38.5	& 42.3	& 94.1	& 31.5	& 97.5	& 68.8	& 61.3	& 54.1	& 61.8\\
    \midrule
    \multirow{1}*{\shortstack{Decoder}}
    & \multirow{1}*{GPT-XL (1.5B)}
        & \multirow{1}*{71.1}
        & \multirow{1}*{75.0}
        & \multirow{1}*{30.4}
        & \multirow{1}*{31.8}
        & \multirow{1}*{37.8}
        & \multirow{1}*{71.0}
        & \multirow{1}*{40.9}
        & \multirow{1}*{87.7}
        & \multirow{1}*{62.5}
        & \multirow{1}*{54.5}
        & \multirow{1}*{50.3}
        & \multirow{1}*{55.7}
    \\
    \bottomrule[1pt]
    \end{tabular}}
    \caption{Ablation study on different backbone models. We experiment with base models of different architectures and scales. ``Enc-Dec'' refers to models that are pretrained in an encoder-decoder manner.}
    \label{tab:ablationbasemodel}
\end{table*}

We also study the effects of using different backbone pretrained models. We experiment with three backbone models of different types, respectively the encoder part of an encoder-decoder model, an encoder model, and a decoder model. Specifically, we use the T5 encoder, DeBERTa \cite{debertav3}, and GPT \cite{radford2018gpt} respectively for these three types. It is noteworthy that though similar in architecture for both T5 encoder and DeBERTa, they are pretrained with different self-supervised language modeling tasks, which in fact leads to huge differences in zero-shot generalization, as we will show in Table~\ref{tab:ablationbasemodel}.

Results of different backbone models are presented in Table \ref{tab:ablationbasemodel}. 
Among all three types of backbone models, the encoder backbone models appear to be the most suitable type of backbone, where both encoder models of two scales respectively achieve the best and the second best results, outperforming all the others by more than 5 points.

Using the same number of parameters (i.e., 1.5B), both DeBERTa-V2 and T5-Encoder significantly outperform GPT-XL, which demonstrates that a bidirectional architecture works better than the unidirectional architecture for the discriminator formulation.
In addition, DeBERTa-V2 outperforms T5-Encoder by 7 points, implying that not only model architecture but also the self-supervised pretraining task determines the ability of UD discrimination. Models pretrained with masked language modeling tasks are more suitable for UD.

The impacts of the architecture and pretraining tasks of backbone models are even larger than the influence of scale, as we also observe that an encoder model with 300M parameters (i.e., DeBERTaV3) achieves much better performance than the T5 encoder and GPT-XL with 1.5B parameters.

\end{document}